\documentclass[10pt,twocolumn,letterpaper]{article}

\usepackage{cvpr}
\usepackage{times}
\usepackage{epsfig}
\usepackage{graphicx}
\usepackage{amsmath}
\usepackage{amssymb}

\usepackage[breaklinks=true,bookmarks=false]{hyperref}

\cvprfinalcopy % *** Uncomment this line for the final submission

\def\cvprPaperID{****} % *** Enter the CVPR Paper ID here

\newif\ifsqueeze
\squeezetrue  % set \squeezetrue or \squeezefalse to enable or disable squeezing
\ifsqueeze

\else

\fi

\begin{document}

%%%%%%%%% TITLE
\title{The iWildCam 2018 Challenge Dataset\thanks{We would like to thank the USGS and NPS for providing us with data. This material is based upon work supported by the National Science Foundation Graduate Research Fellowship Program under Grant No. 1745301. Any opinions, findings, and conclusions or recommendations expressed in this material are those of the author(s) and do not necessarily reflect the views of the National Science Foundation.}}

\author{Sara Beery, Oisin Mac Aodha, Grant van Horn, Pietro Perona \\
California Institute of Technology\\
1200 E California Blvd., Pasadena, CA 91125\\
{\tt\small sbeery@caltech.edu, macaodha@caltech.edu, gvanhorn@caltech.edu, perona@caltech.edu}
}

\maketitle
%\thispagestyle{empty}

%%%%%%%%% ABSTRACT
\begin{abstract}
Camera traps are a valuable tool for studying biodiversity, but research using this data is limited by the speed of human annotation. With the vast amounts of data now available it is imperative that we develop automatic solutions for annotating camera trap data in order to allow this research to scale. A promising approach is based on deep networks trained on human-annotated images~\cite{norouzzadeh2017automatically}.  We provide a challenge dataset to explore whether such solutions generalize to novel locations, since systems that are trained once and may be deployed to operate automatically in new locations would be most useful. 
\end{abstract}
\section{Introduction}
As the planet changes due to urbanization and climate change, biodiversity worldwide is in decline. We are currently witnessing an estimated rate of species loss that is up to 200 times greater than historical rates \cite{CBD2017}. Monitoring biodiversity quantitatively can help us understand the connections between species decline and pollution, exploitation, urbanization, global warming, and conservation policy. %\cite{butchart2010global,cardinale2012biodiversity}.
\begin{figure}
\begin{minipage}[b]{.24\linewidth}
  \centering
  \centerline{\includegraphics[width=2.2cm]{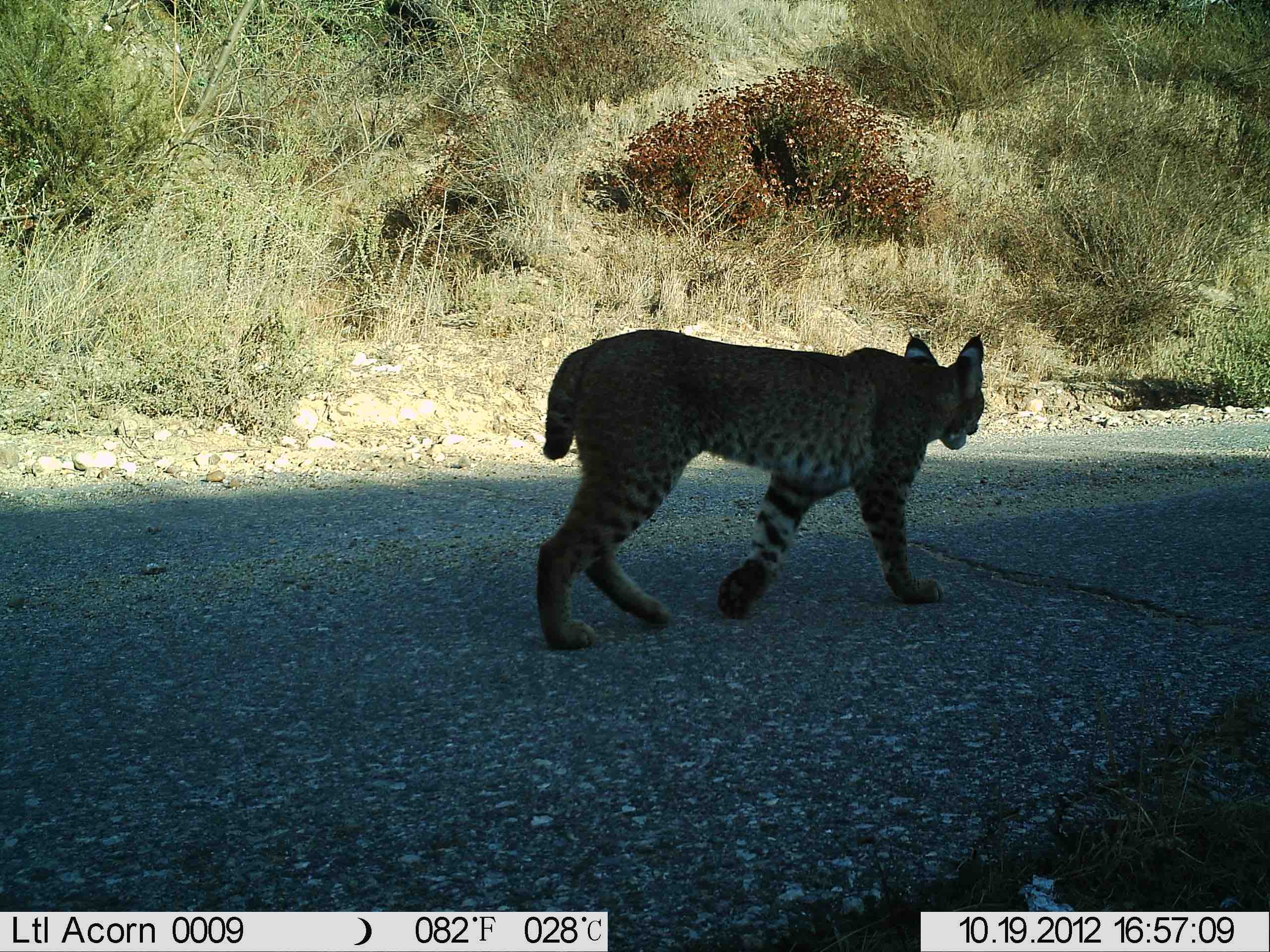}}
  \vspace{.05cm}
\end{minipage}
\hfill
\begin{minipage}[b]{0.24\linewidth}
  \centering
  \centerline{\includegraphics[width=2.2cm]{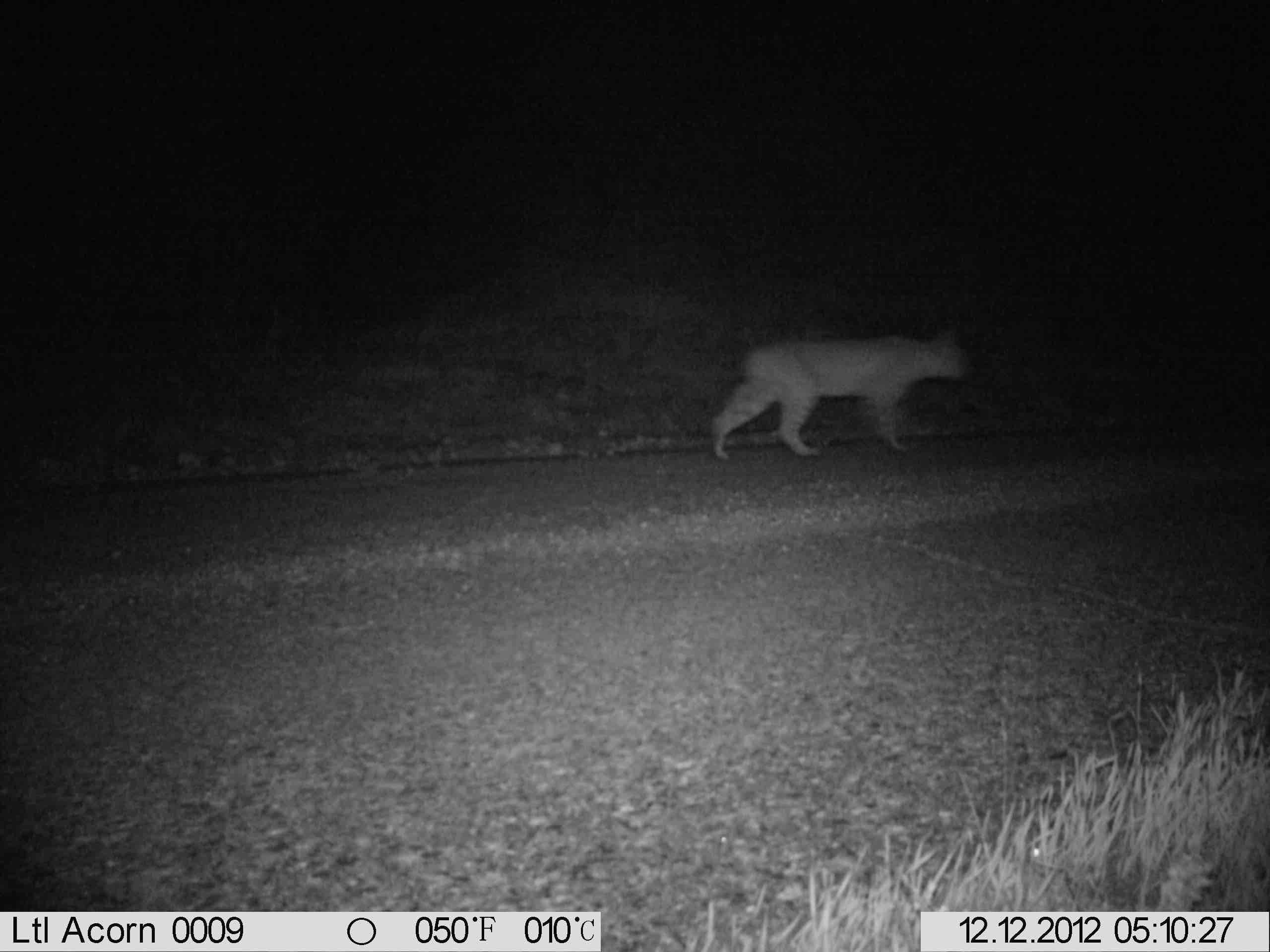}}
\vspace{.05cm}

\end{minipage}
\hfill
\begin{minipage}[b]{.24\linewidth}
  \centering
  \centerline{\includegraphics[width=2.2cm]{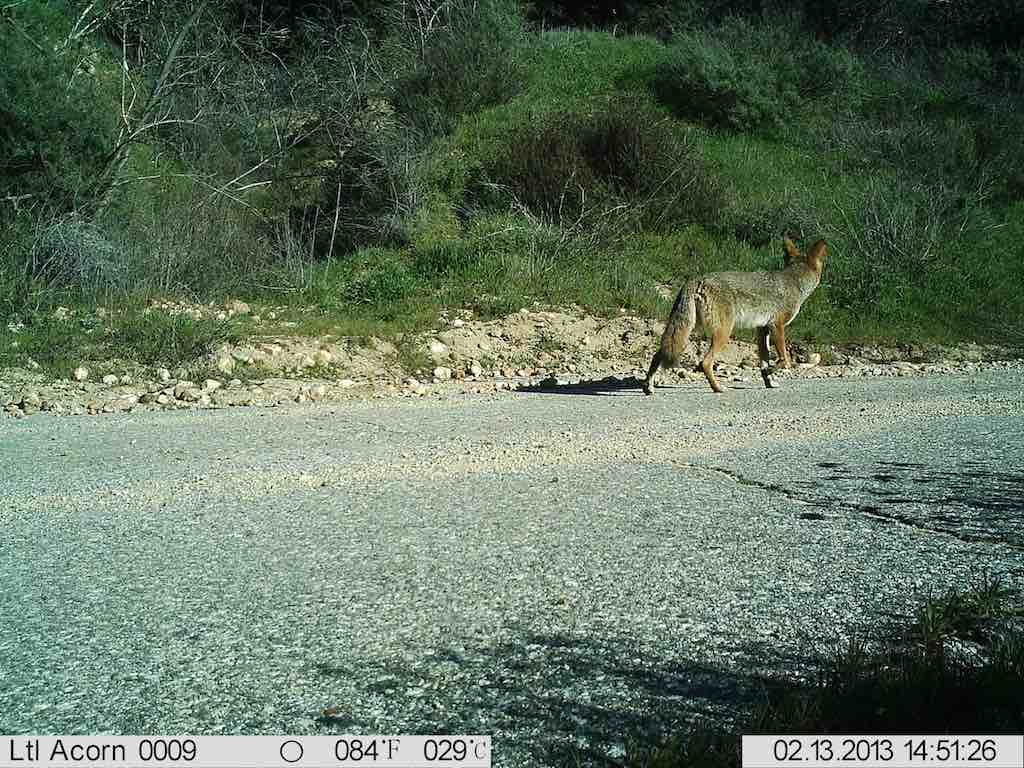}}
\vspace{.05cm}

\end{minipage}
\hfill
\begin{minipage}[b]{.24\linewidth}
  \centering
  \centerline{\includegraphics[width=2.2cm]{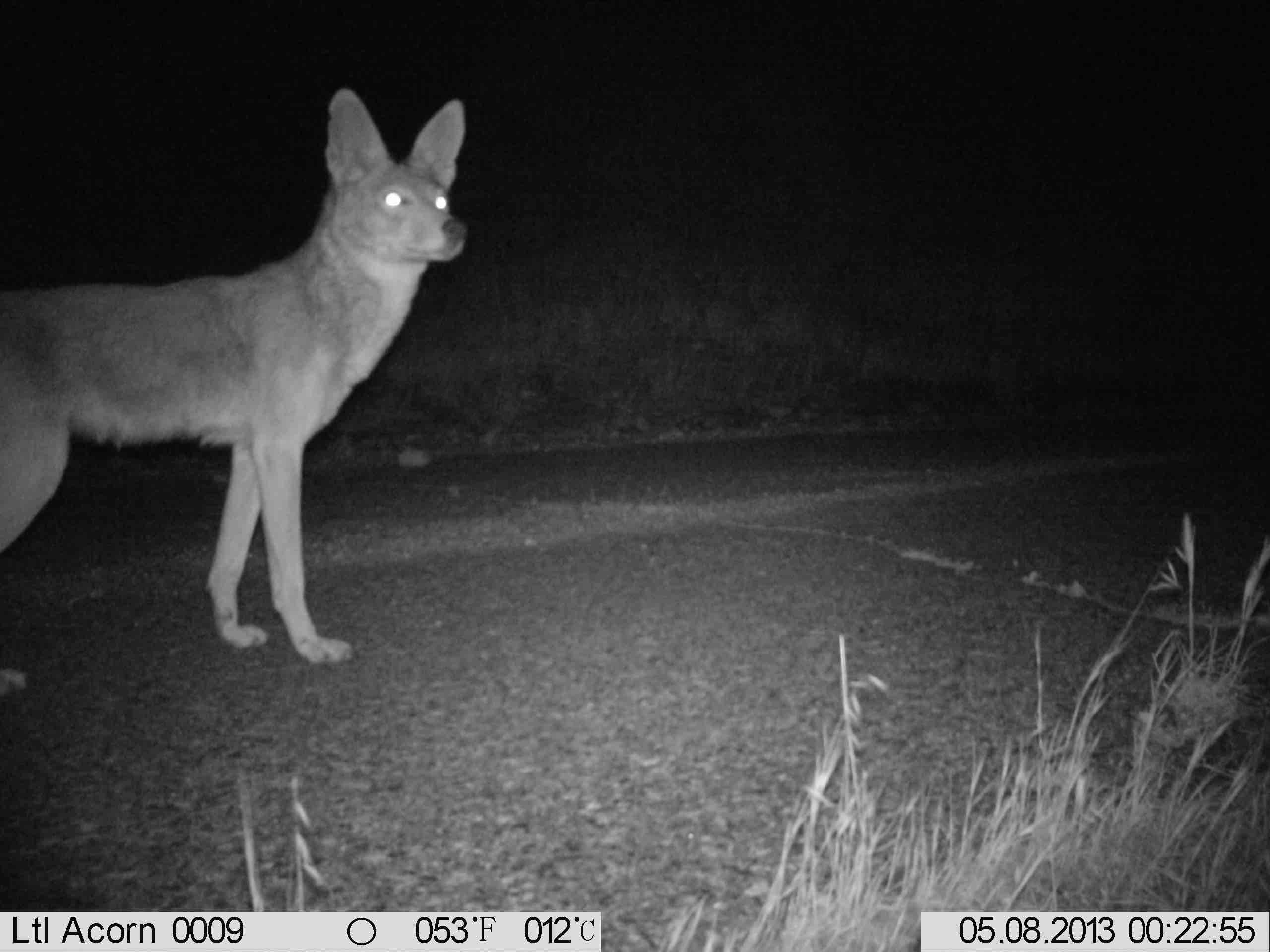}}
\vspace{.05cm}

\end{minipage}

\begin{minipage}[b]{0.24\linewidth}
  \centering
  \centerline{\includegraphics[width=2.2cm]{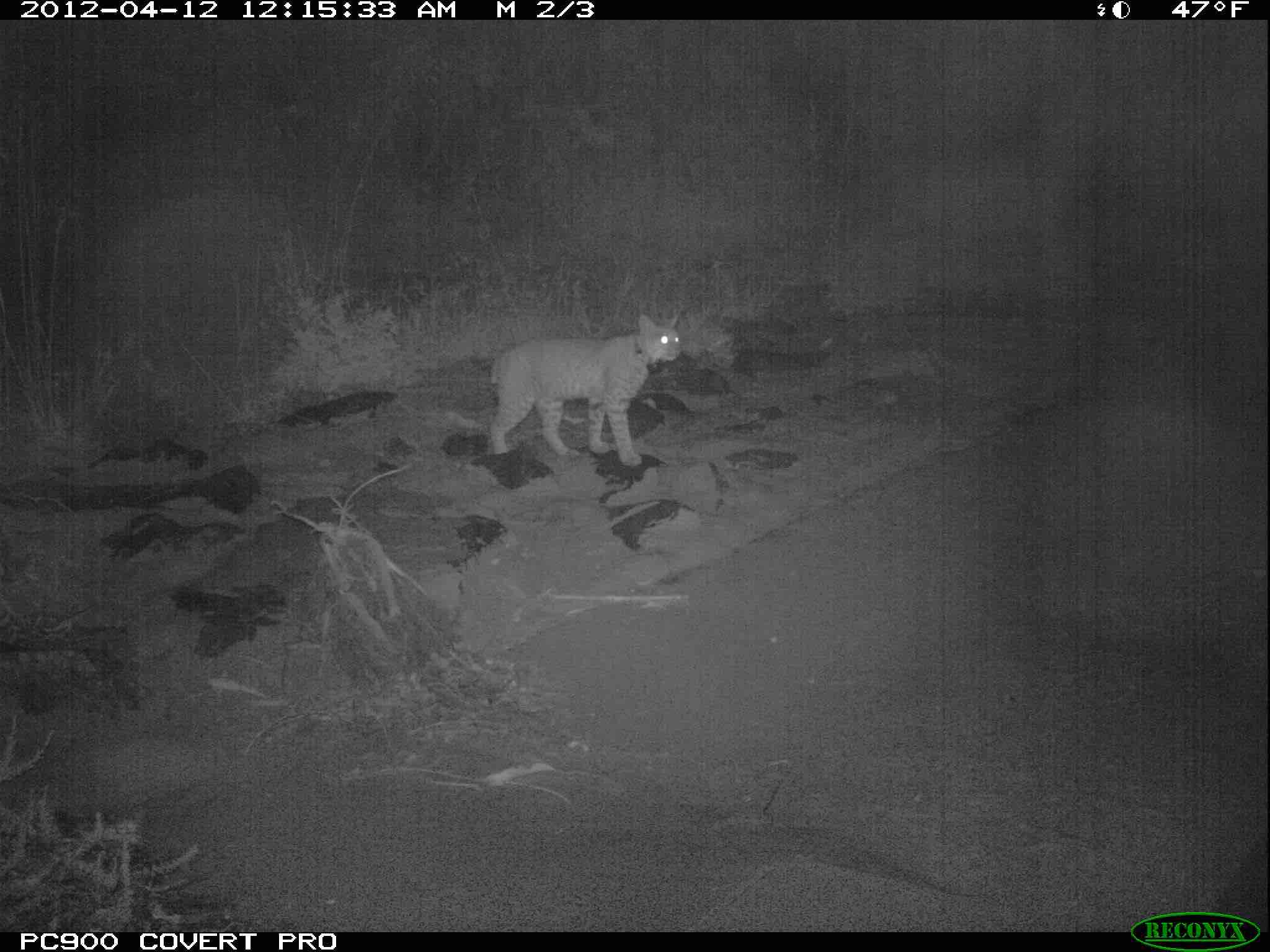}}
\vspace{.05cm}

\end{minipage}
\hfill
\begin{minipage}[b]{.24\linewidth}
  \centering
  \centerline{\includegraphics[width=2.2cm]{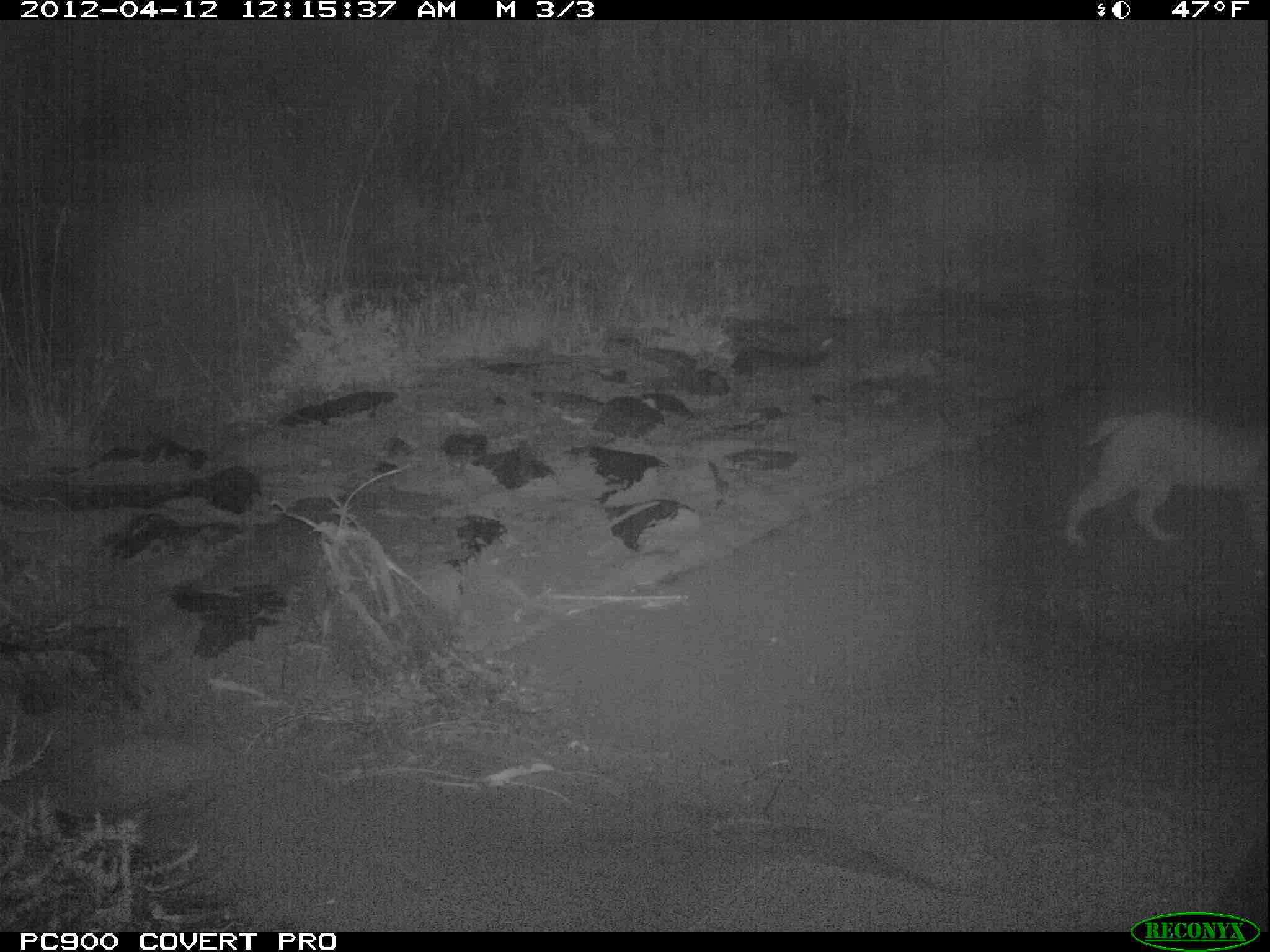}}
\vspace{.05cm}

\end{minipage}
\hfill
\begin{minipage}[b]{0.24\linewidth}
  \centering
  \centerline{\includegraphics[width=2.2cm]{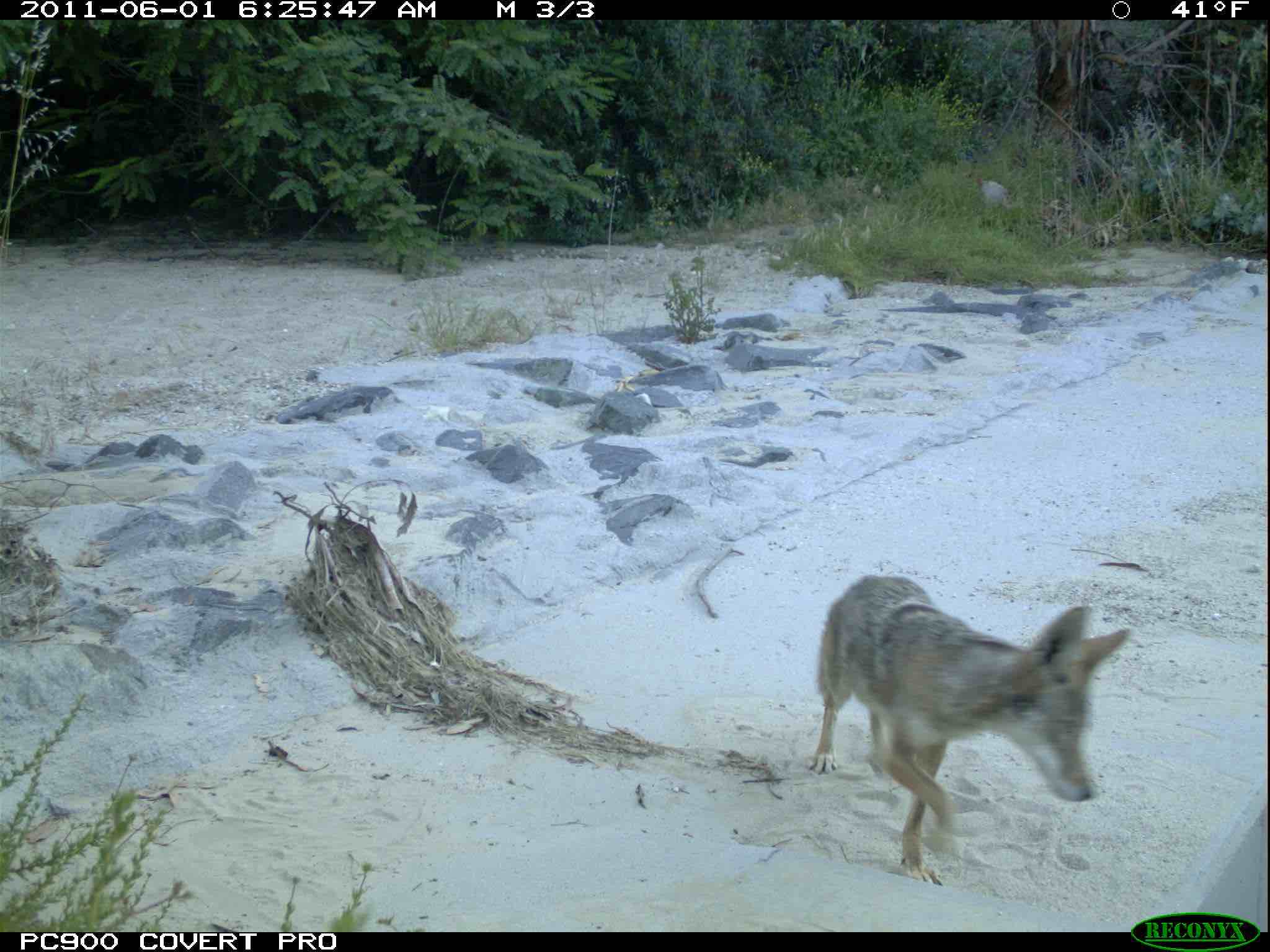}}
\vspace{.05cm}

\end{minipage}
\hfill
\begin{minipage}[b]{.24\linewidth}
  \centering
  \centerline{\includegraphics[width=2.2cm]{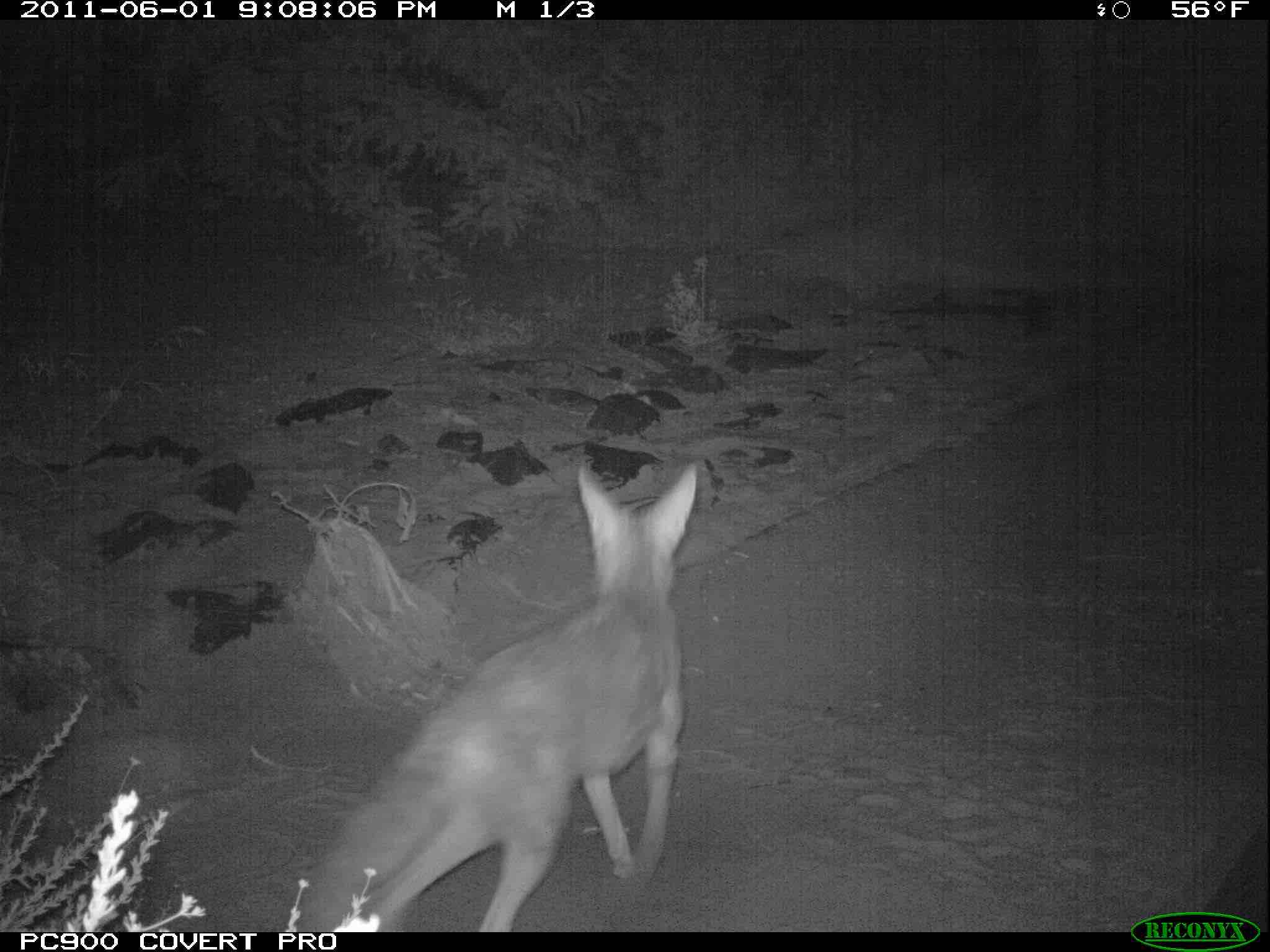}}
\vspace{.05cm}

\end{minipage}

\begin{minipage}[b]{0.24\linewidth}
  \centering
  \centerline{\includegraphics[width=2.2cm]{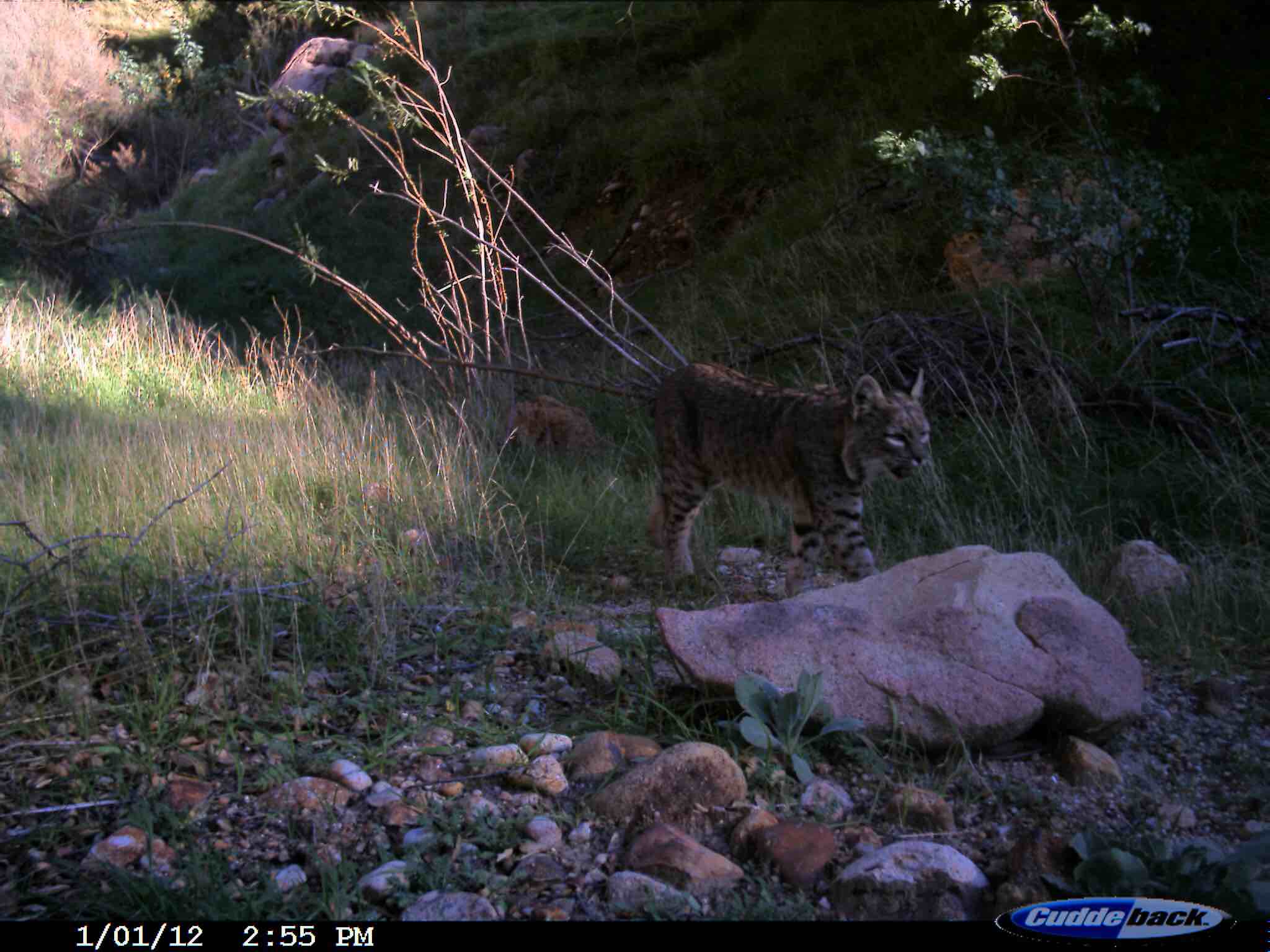}}
%\vspace{.05cm}

\end{minipage}
\hfill
\begin{minipage}[b]{.24\linewidth}
  \centering
  \centerline{\includegraphics[width=2.2cm]{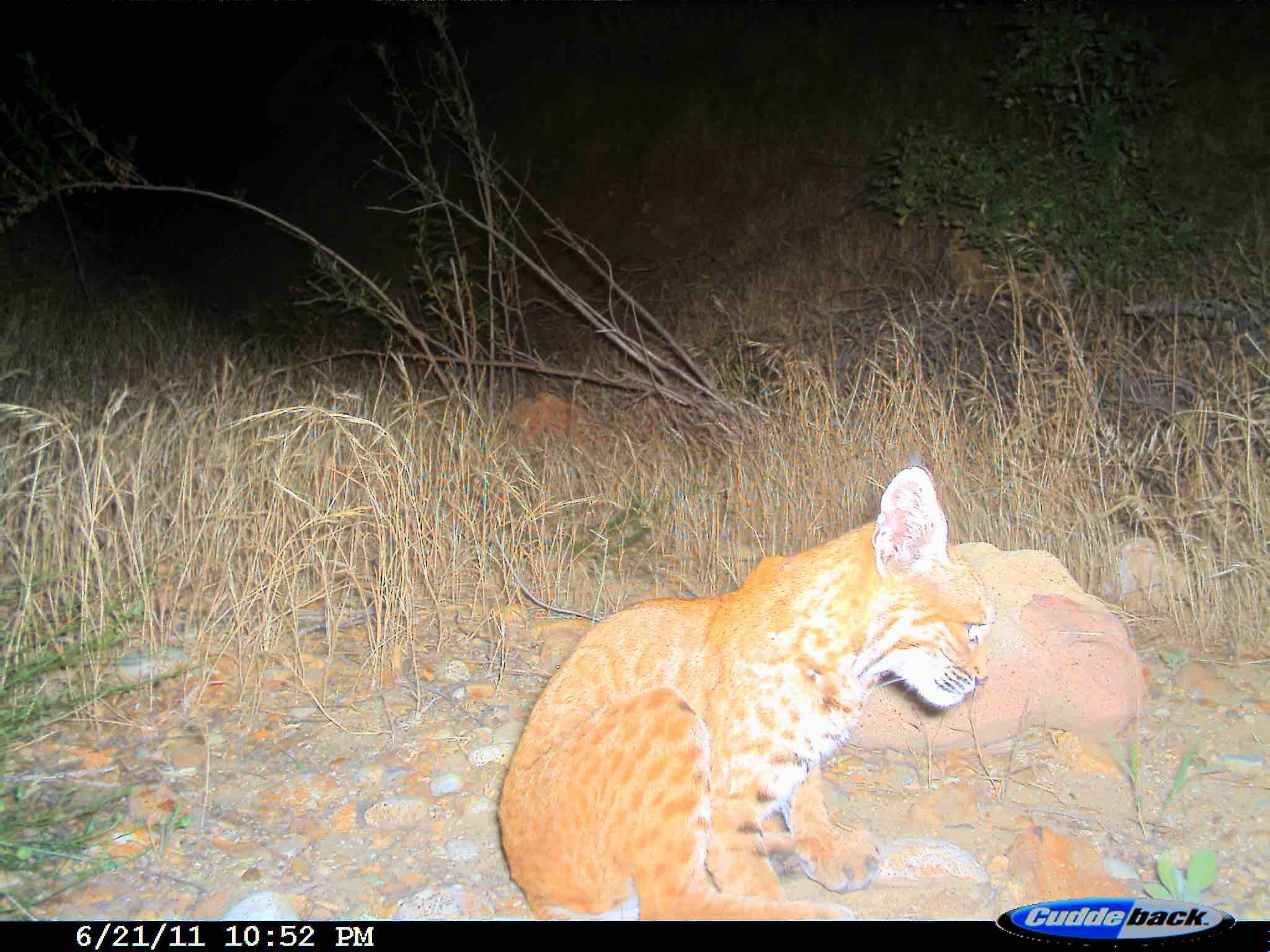}}
%\vspace{.05cm}

\end{minipage}
\hfill
\begin{minipage}[b]{0.24\linewidth}
  \centering
  \centerline{\includegraphics[width=2.2cm]{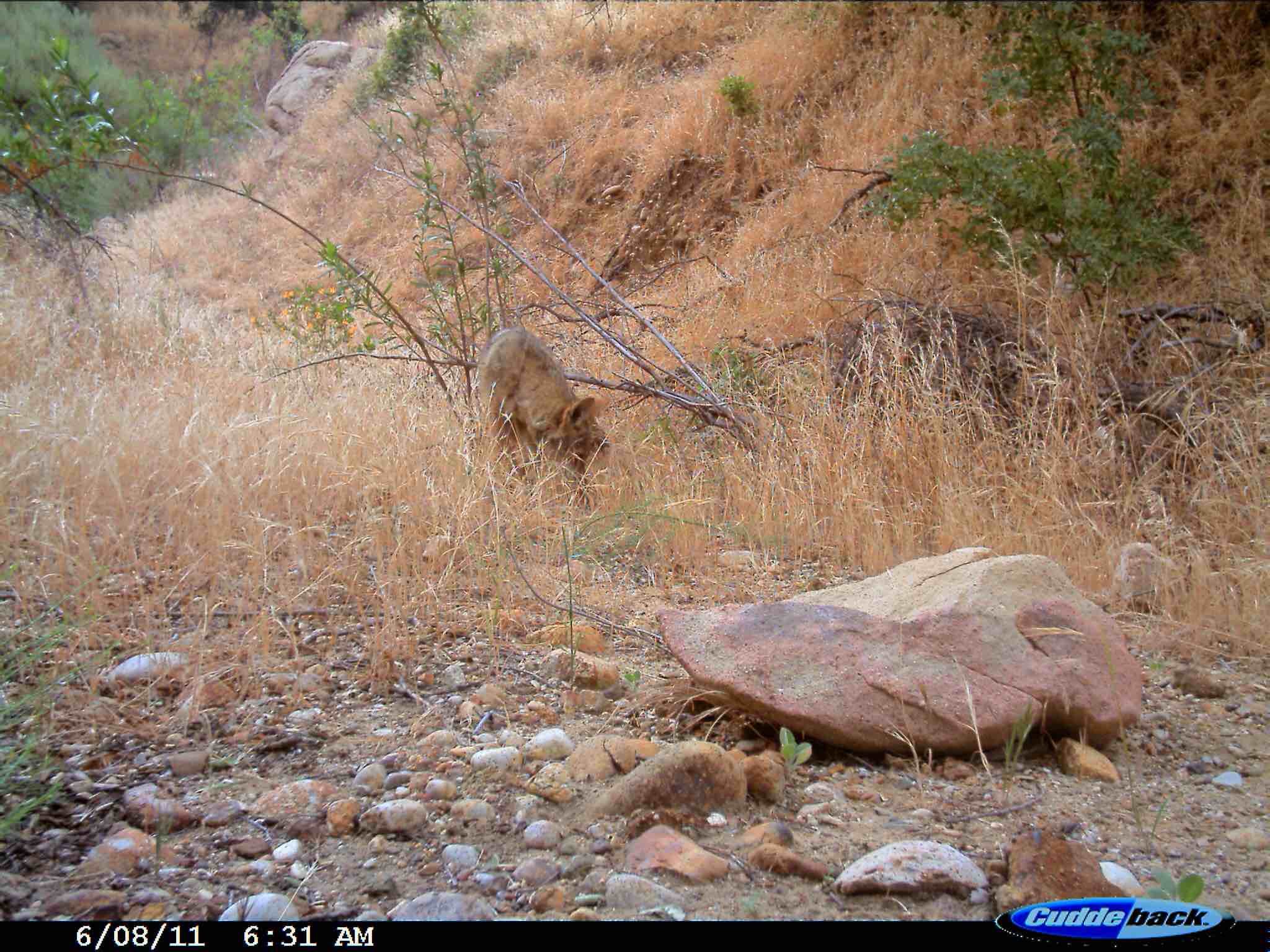}}
%\vspace{.05cm}

\end{minipage}
\hfill
\begin{minipage}[b]{.24\linewidth}
  \centering
  \centerline{\includegraphics[width=2.2cm]{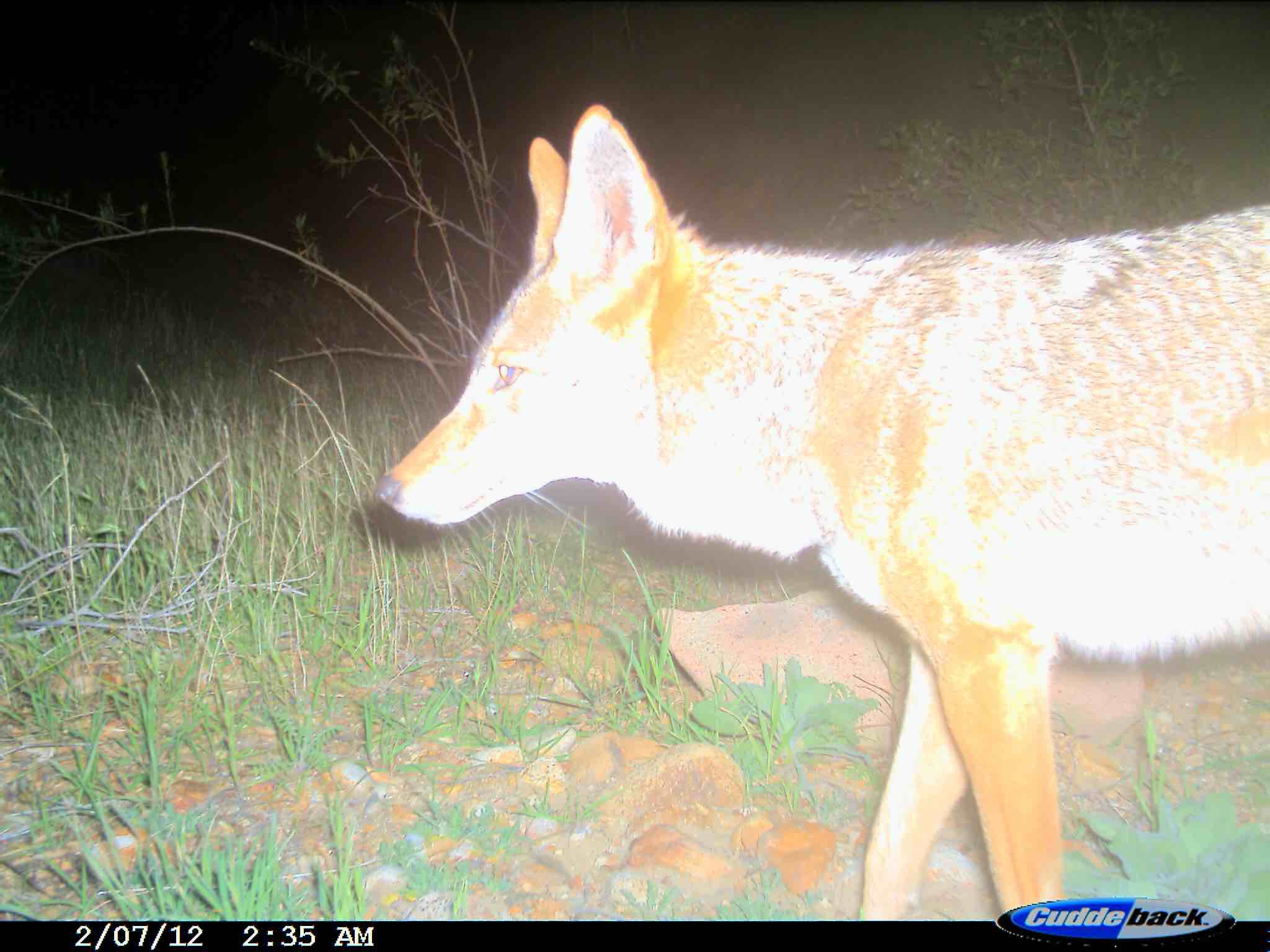}}
%\vspace{.05cm}
\end{minipage}
\caption{{\bf Camera trap images from three different locations.} Each row is a different location and a different camera type. The first two cameras use IR, while the third row used white flash.  The first two columns are bobcats, the next two columns are coyotes.}
\label{fig:camtrap_ims}
\end{figure}
We collaborate with researchers who study the effect of these factors on wild animal populations by monitoring changes in species diversity, population density, and behavioral patterns. In the past, much of this research was done via field studies, where biologists and zoologists would study animals in the wild by direct observation. However, once cameras became relatively inexpensive and easy to use, many scientists turned to camera traps as an efficient, cost-effective, non-invasive method to collect experimental data. Camera traps are heat- or motion-activated cameras placed in the wild to monitor and investigate animal populations and behavior. They are used to look for endangered and threatened species, to help identify important habitats, to monitor sites of interest, to create an index for population changes over time, and to analyze wildlife activity patterns. 

At present, images are annotated by hand, and the time required to sort images severely limits data scale and research productivity.  Our collaborators estimate that they can annotate around 3 images/minute, and spend close to 500 hours per project on data annotation. Annotation of camera trap photos is not only time consuming, but it is also challenging. Because the images are taken automatically based on a triggered sensor, there is no guarantee that the animal will be centered, focused, well-lit, or an appropriate scale (they can be either very close or very far from the camera, each causing its own problems). Further, up to 70\% of the photos at any given location are triggered by something other than an animal, such as wind in the trees, a passing car, or a hiker. Going through these photos manually is not productive from a research perspective. 

Camera traps are not new to the computer vision community \cite{ren2013ensemble,yu2013automated,wilber2013animal,chen2014deep,lin2014foreground,swanson2015snapshot,zhang2015coupled,zhang2016animal,miguel2016finding,giraldo2017camera,yousif2017fast,villa2017towards,norouzzadeh2017automatically}. 
However, each of the previous methods have used the same camera locations for both training and testing the performance of an automated system. If we wish to build systems that are trained once to detect and classify animals, and then deployed to new locations without further training, we must measure the ability of machine learning and computer vision to {\em generalize} to new environments. This dataset is the first to focus on the need for generalization in automated solutions for camera trap data.

\section{The iWildCam Dataset}
\begin{figure}
\centering
\includegraphics[height=5cm]{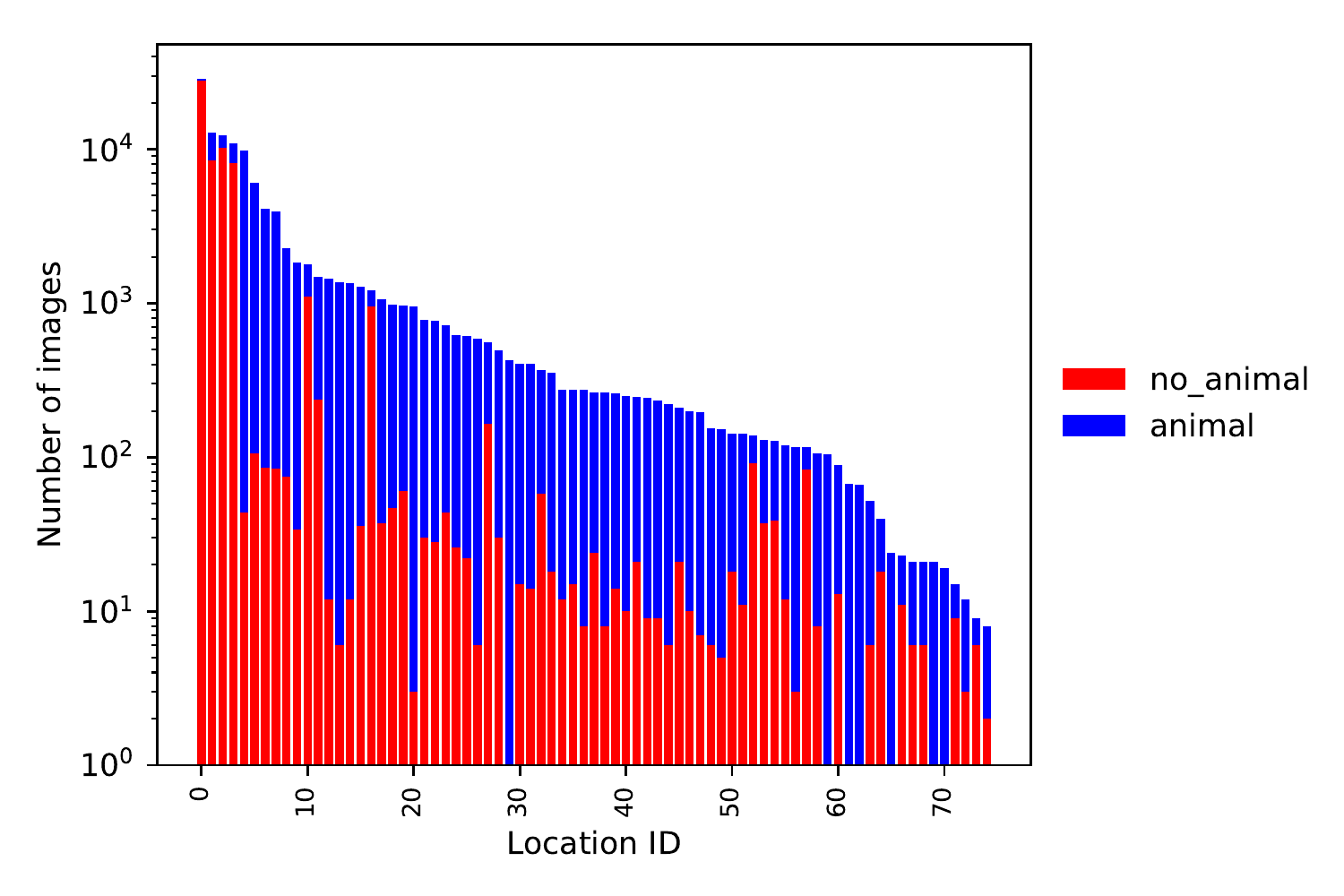}  

\caption{Number of annotations for each location, over the two classes. The distribution images per location is long-tailed, and each location has a different and peculiar class distribution. }
\label{fig:annotPerLoc}
\end{figure}

All images in our dataset come from the American Southwest. By limiting the geographic region, the flora and fauna seen across the locations remain consistent. The current task is not to deal with entirely new regions or species, but instead to be able to recognize the same species of animals in the same region with a different camera background. In the future we plan to extend this dataset to include other regions, in order to tackle the challenges of both recognizing animals in new regions, and to the open-set problem of recognizing species of animals that have never before been seen. Examples of data from different locations can be seen in Fig.~\ref{fig:camtrap_ims}. 
Our dataset consists of $292,732$ images across $143$ locations, each labeled as either containing an animal, or as empty. See Fig.~\ref{fig:annotPerLoc} for the distribution of classes and images across locations. We do not filter the stream of images collected by the traps, rather this is the same data that a human biologist currently sifts through. Therefore the data is unbalanced in the number of images per location, distribution of species per location, and distribution of species overall (see Fig.~\ref{fig:annotPerLoc}). The class of each image was provided by expert biologists from the NPS and USGS. Due to different annotation styles and challenging images, we approximate that the dataset contains up to 5\% annotation error.

\subsection{Data Challenges}

\begin{figure}
\begin{minipage}[b]{.3\linewidth}
  \centering
  \centerline{\includegraphics[width=2.25cm]{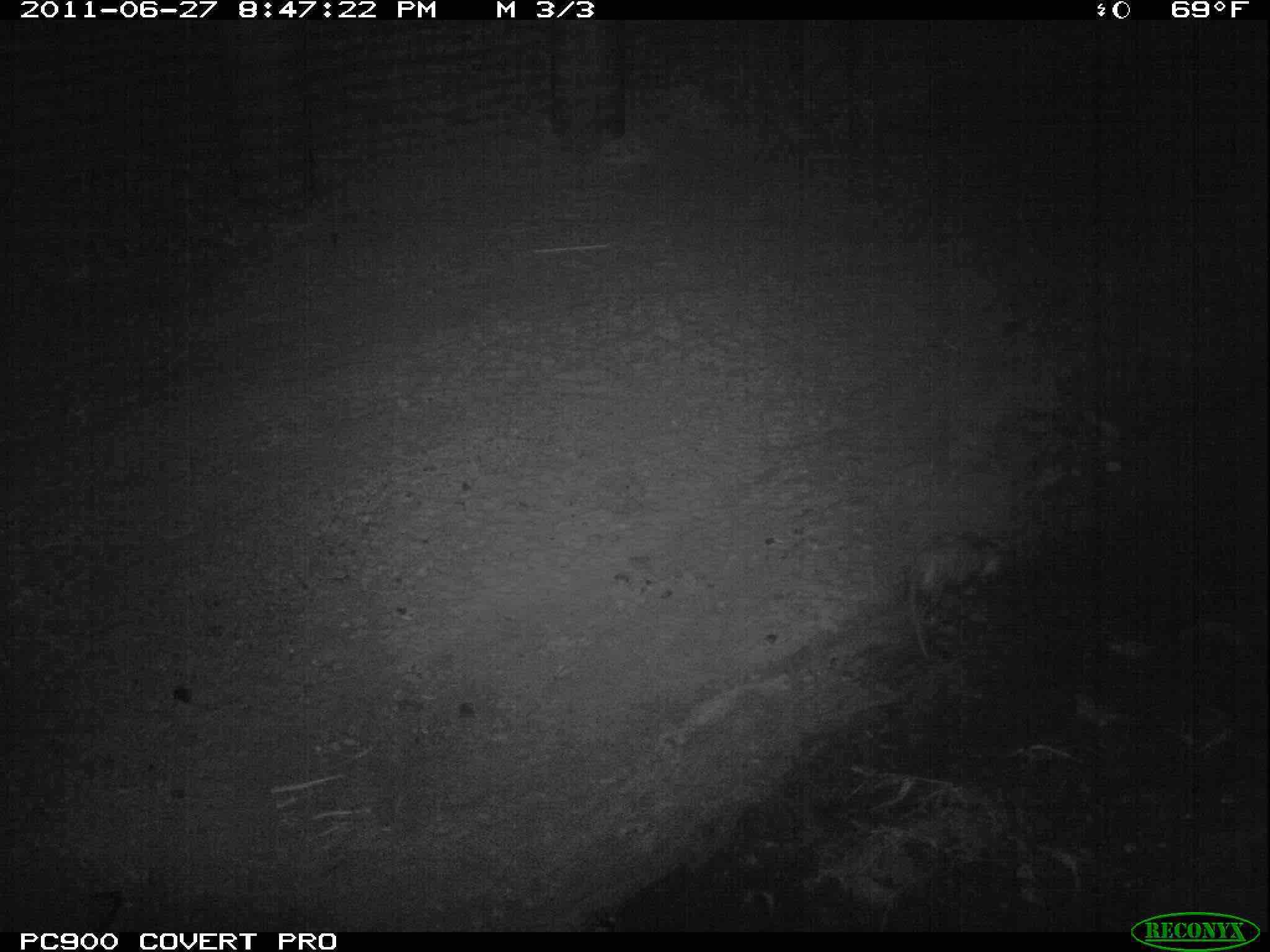}}
  \vspace{.05cm}
%   \centerline{(1) Illumination}\medskip
\end{minipage}
\hfill
\begin{minipage}[b]{0.3\linewidth}
  \centering
  \centerline{\includegraphics[width=2.25cm]{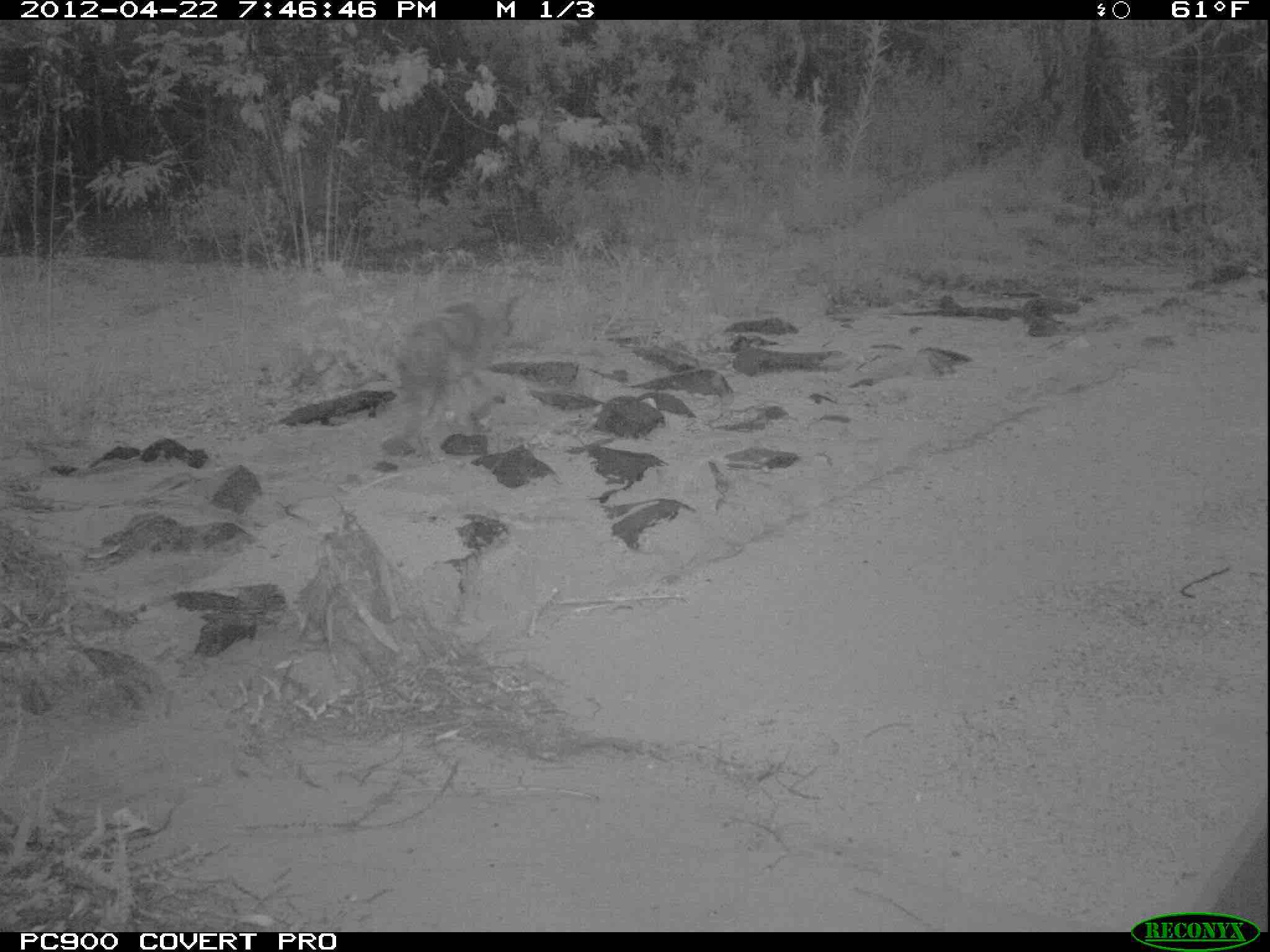}}
  \vspace{.05cm}
%   \centerline{(2) Blur}\medskip
\end{minipage}
\hfill
\begin{minipage}[b]{.3\linewidth}
  \centering
  \centerline{\includegraphics[width=2.25cm]{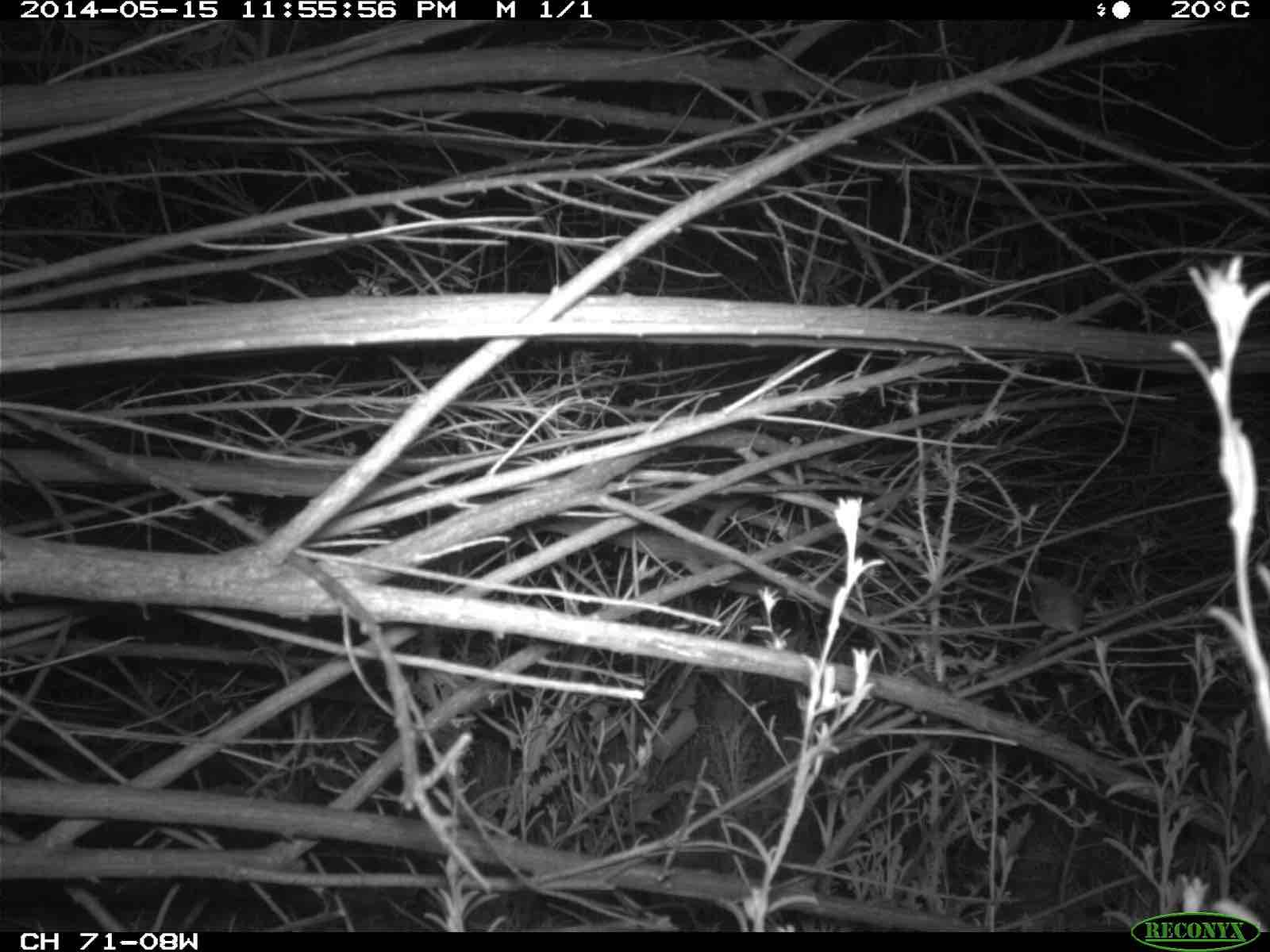}}
  \vspace{.05cm}
%   \centerline{(3) ROI Size}\medskip
 \end{minipage}
\begin{minipage}[b]{0.3\linewidth}
  \centering
  \centerline{\includegraphics[width=2.25cm]{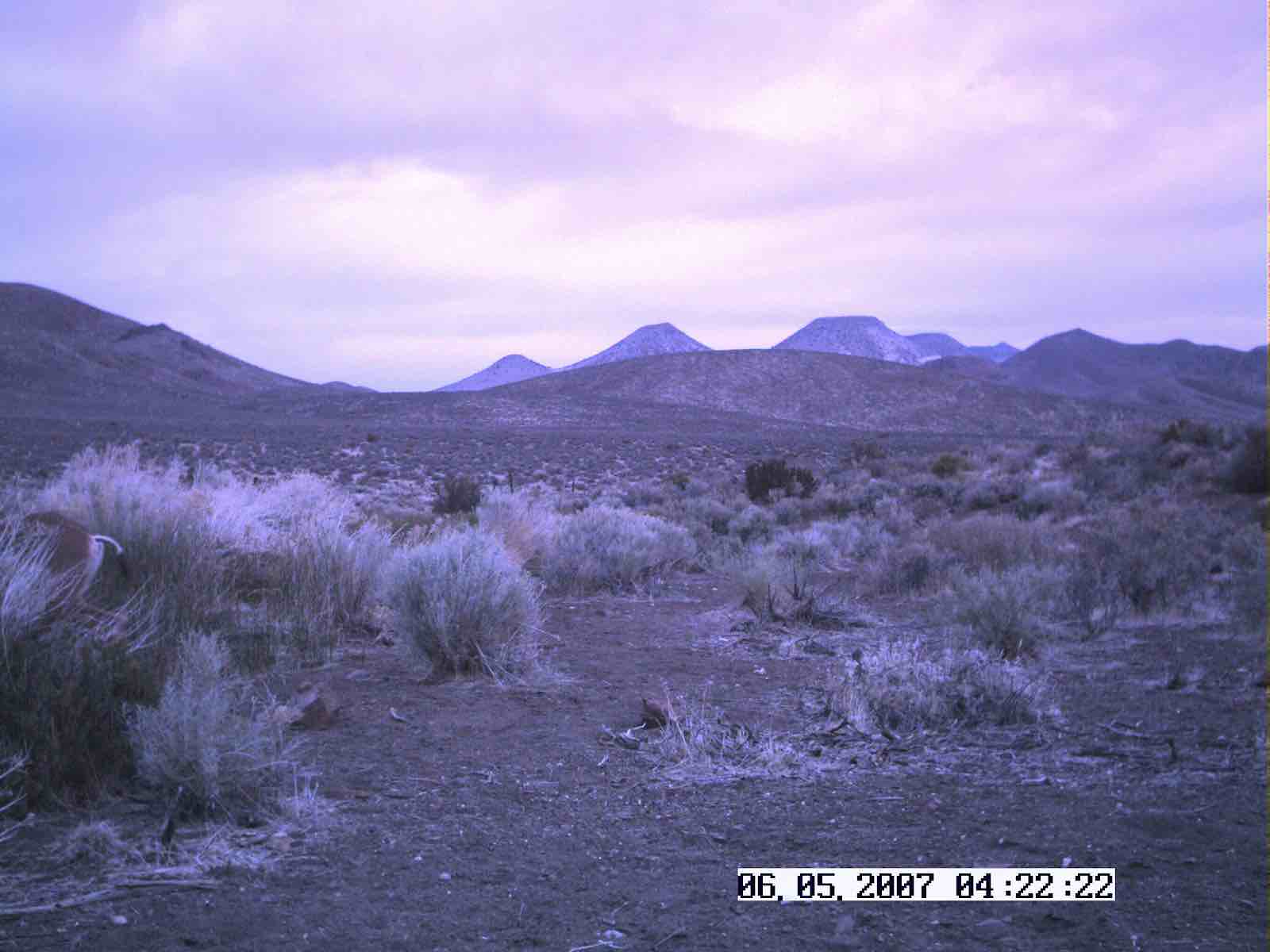}}
%  \vspace{1.5cm}
%   \centerline{(4) Occlusion}\medskip
\end{minipage}
\hfill
\begin{minipage}[b]{.3\linewidth}
  \centering
  \centerline{\includegraphics[width=2.25cm]{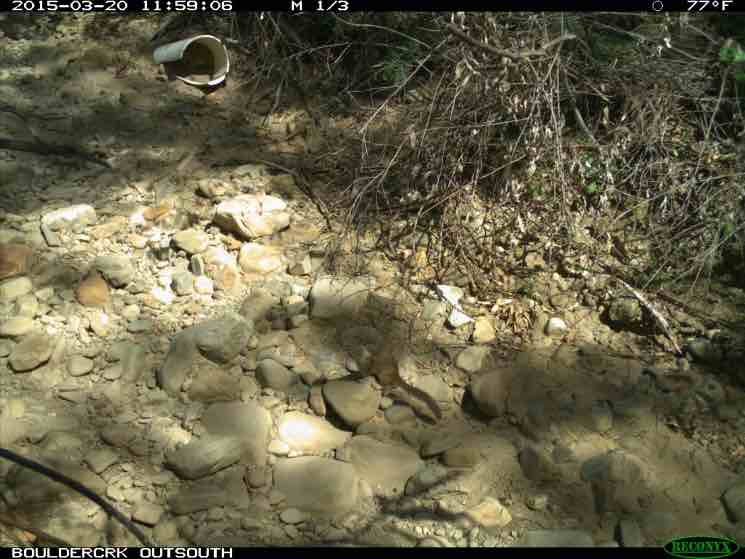}}
%  \vspace{1.5cm}
%   \centerline{(5) Camouflage}\medskip
\end{minipage}
\hfill
\begin{minipage}[b]{0.3\linewidth}
  \centering
  \centerline{\includegraphics[width=2.25cm]{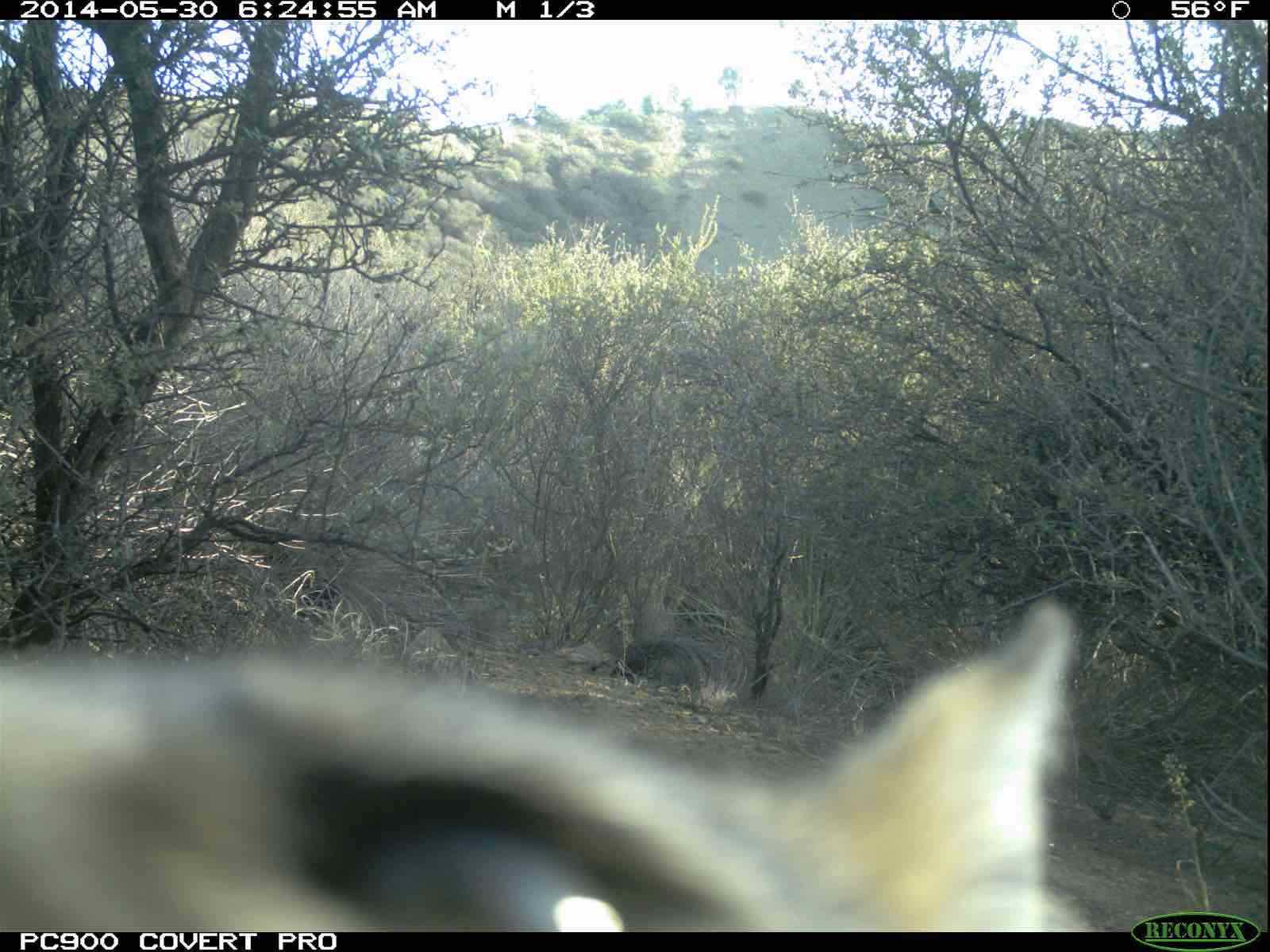}}
%  \vspace{1.5cm}
%   \centerline{(6) Perspective}\medskip
\end{minipage}
\caption{\textbf{Common data challenges}: (1) {\bf Illumination}: Animals are not always salient. (2) {\bf Motion blur}: common with poor illumination at night. (3) {\bf Size of the region of interest} (ROI): Animals can be small or far from the camera. (4) {\bf Occlusion}: e.g. by bushes or rocks. (5) {\bf Camouflage}: decreases saliency in animals' natural habitat. (6) {\bf Perspective}: Animals can be close to the camera, resulting in partial views of the body.}
\label{fig:challenging_ims}
\end{figure}

The animals in the images can be challenging to detect, even for humans. We find six main nuisance factors inherent to camera trap data (Fig.~\ref{fig:challenging_ims}). When an image is too difficult to classify on its own, biologists will often refer to an easier image in the same sequence and then track motion by flipping between sequence frames in order to generate a label for each frame (\eg\ is the animal still present or has it gone off the image plane?). This implies that sequence information is a valuable tool in difficult cases. 

\subsection{Data Split and Baseline}
From our pool of $143$ locations, we selected $70$ locations at random to use as training data. We selected $10\%$ of the data from our training locations and 5 random new locations to use as validation data. The remaining $68$ locations are used as test data. This gives us $149,359$ training images, $17,784$ validation and $125,589$ test images. 

We trained a baseline model using the InceptionV3 architecture, pretrained on ImageNet, with an initial learning rate of 0.0045, rmsprop with a momentum of 0.9, and a square input resolution of 299. We employed random cropping (containing most of the region), horizontal flipping, and random color distortion as data augmentation. This baseline achieved 74.1\% accuracy on the test set.

\section{The iWildCam Challenge 2018}
\begin{figure}
\centering
\includegraphics[height=5cm]{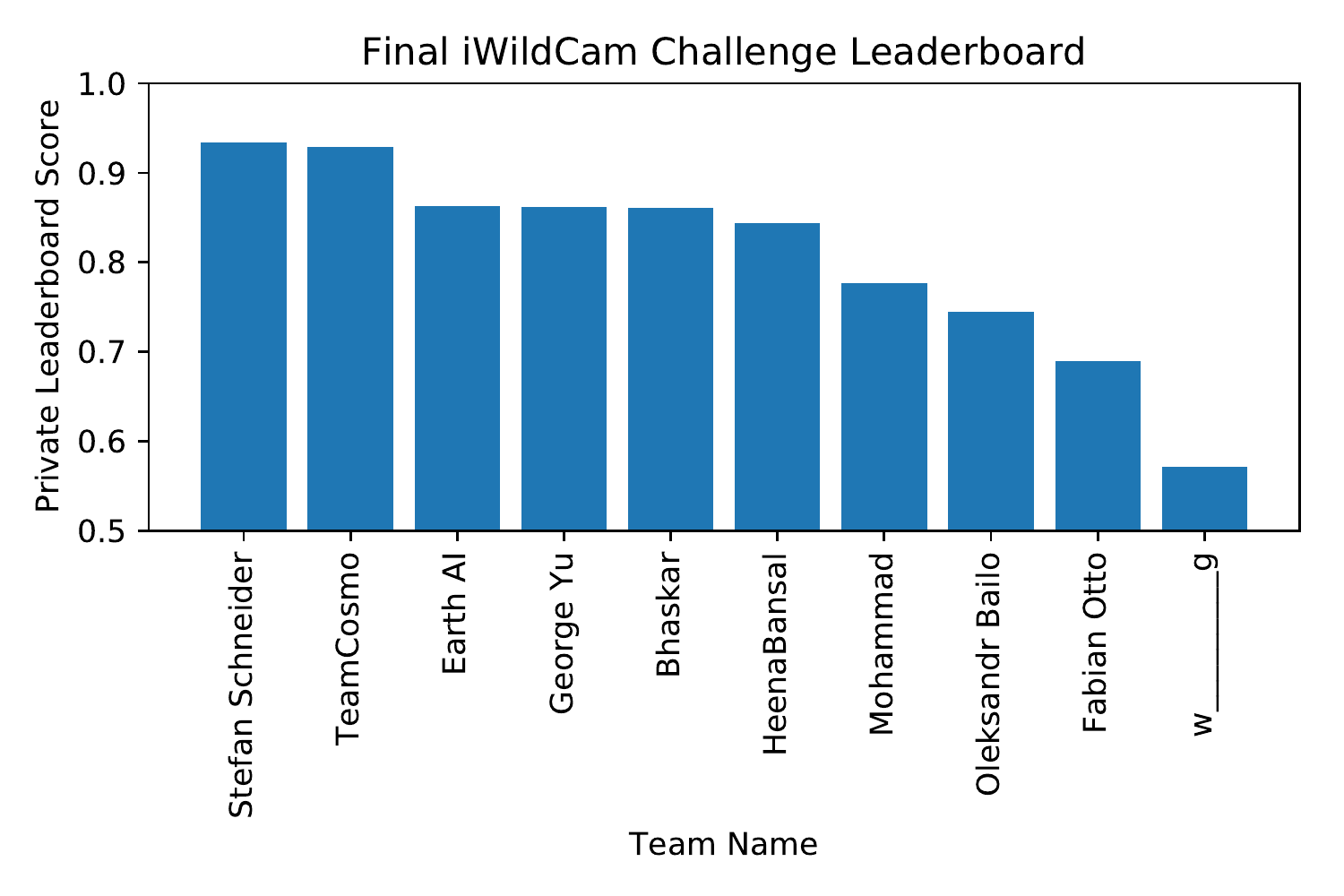}
\caption{The final private leaderboard from the iWildCam Challenge 2018. These results show accuracies on the 50\% held-out private test data randomly selected by Kaggle.}
\label{fig:winners}
\end{figure}

The iWildCam Challenge 2018 was conducted through Kaggle as part of FGVC5 at CVPR18 and had 10 participating teams.  The final leaderboard from the held-out private test data can be seen in Fig. \ref{fig:winners}. The winning method by Stefan Schneider achieved an accuracy of 93.431\%. It consisted of an ensemble of 5 models considering 5 different image sizes (50, 75, 100, 125, 150), all based on the VGG16 architecture. The models were trained from scratch using the Adam optimizer and data augmentation tools were used to randomly flip the images along the horizontal axis and add a range of blurring during training. Stefan considered a variety of models including AlexNet, GoogLeNet, DenseNet, ResNet and his own personal networks in the ensemble but found VGG16 outperformed all of them. He also considered a domain adaption model in an attempt to remove associations of location from the model but found this did not improve overall performance.

\section{Conclusions}
Camera traps provide a unique experimental setup that allow us to explore the generalization of models while controlling for many nuisance factors. This dataset is the first to propose using camera traps to study generalization, and any forward progress made will have a direct impact on the scalability of biodiversity research.

In subsequent years, we plan to extend the iWildCam Challenge by providing per-species class annotations, bounding boxes, and image sequence information, and by adding new locations, both from the American Southwest and from new regions worldwide. Per-species classes will allow us to use the dataset to explore fine-grained classification challenges for the natural world in a challenging setting. Bounding boxes and image sequence information will allow us to explore how animal movement through a scene can be used to improve current results. Drastic landscape and vegetation changes will allow us to investigate generalization in an even more challenging setting. By testing on regions from across the globe, we can investigate the open-set problem of detecting animal species not seen during training.

We have already curated and collected macro-class annotations (\eg "bird" or "rodent") for all images in the current dataset, as well as bounding box annotations for a subset of $57,868$ images covering 20 locations. These additional annotations will be publicly released in Summer 2018.

{\small
\bibliographystyle{ieee}
\bibliography{main}
}

\end{document}